\documentclass[conference]{IEEEtran}
\IEEEoverridecommandlockouts
\usepackage{cite}
\usepackage{amsmath,amssymb,amsfonts}
\usepackage{algorithm}
\usepackage[noend]{algpseudocode}
\usepackage[hidelinks]{hyperref}
\usepackage{graphicx}
\usepackage{textcomp}
\usepackage{xcolor}
\usepackage{subfigure}
\usepackage{booktabs}
\usepackage{xspace}
\usepackage{eso-pic}

\newcommand{\ours}{\textcolor{black}{T3S}\xspace}
\newcommand{\fullours}{\textcolor{black}{\textbf{T}ask-\textbf{S}pecific feature \textbf{S}elector and \textbf{S}cheduler (T3S)}\xspace}
\newcommand{\shared}{\textcolor{black}{globally shared network}\xspace}
\newcommand{\selectors}{\textcolor{black}{feature selectors}\xspace}
\newcommand{\selector}{\textcolor{black}{feature selector}\xspace}

\def\BibTeX{{\rm B\kern-.05em{\sc i\kern-.025em b}\kern-.08em
    T\kern-.1667em\lower.7ex\hbox{E}\kern-.125emX}}
\begin{document}

\AddToShipoutPictureBG*{%
  \AtPageLowerLeft{%
    \hspace*{\dimexpr1in+\oddsidemargin\relax}%
    \raisebox{0.25in}{%
      \parbox[b]{\textwidth}{%
        \fontsize{7}{8}\selectfont
        \textbf{Author's accepted manuscript.} Published version: Y. Yu, T. Yang, Y. Lv,
        Y. Zheng, and J. Hao, ``T3S: Improving Multi-Task Reinforcement Learning
        with Task-Specific Feature Selector and Scheduler,'' \textit{2023
        International Joint Conference on Neural Networks (IJCNN)}, 2023,
        pp.~1--8. DOI: \href{https://doi.org/10.1109/IJCNN54540.2023.10191536}{10.1109/IJCNN54540.2023.10191536}.\par
        \textcopyright\ 2023 IEEE. Personal use of this material is permitted.
        Permission from IEEE must be obtained for all other uses, in any current
        or future media, including reprinting/republishing this material for
        advertising or promotional purposes, creating new collective works, for
        resale or redistribution to servers or lists, or reuse of any copyrighted
        component of this work in other works.
      }%
    }%
  }%
}

\title{T3S: Improving Multi-Task Reinforcement Learning with Task-Specific Feature Selector and Scheduler\\
\thanks{$^*$ Corresponding author.}
}

\author{\IEEEauthorblockN{Yuanqiang Yu}
\IEEEauthorblockA{\textit{College of Intelligence and Computing} \\
\textit{Tianjin University}\\
Tianjin, China \\
yuyuanqiang@tju.edu.cn}
\and
\IEEEauthorblockN{Tianpei Yang$^*$}
\IEEEauthorblockA{\textit{Department of Computing Science} \\
\textit{University of Alberta and Alberta Machine Intelligence Institute}\\
Edmonton, Canada \\
tpyang@tju.edu.cn}
\and
\IEEEauthorblockN{Yongliang Lv}
\IEEEauthorblockA{\textit{College of Intelligence and Computing} \\
\textit{Tianjin University}\\
Tianjin, China \\
lvyongliang@tju.edu.cn}
\and
\IEEEauthorblockN{Yan Zheng}
\IEEEauthorblockA{\textit{College of Intelligence and Computing} \\
\textit{Tianjin University}\\
Tianjin, China \\
yanzheng@tju.edu.cn}
\and
\IEEEauthorblockN{Jianye Hao$^*$}
\IEEEauthorblockA{\textit{College of Intelligence and Computing} \\
\textit{Tianjin University}\\
Tianjin, China \\
jianye.hao@tju.edu.cn}
}

\maketitle

\begin{abstract}

Multi-task reinforcement learning (MTRL) is a technique to train multiple tasks simultaneously, where previous works usually train a single model to solve different tasks by sharing parameters across various tasks.
However, these methods are faced with inter-task interference since what parameters should be shared across tasks is not addressed, dramatically reducing learning efficiency.
To solve these problems, we propose a novel MTRL framework called \fullours, which consists of two components: a \selector and a task scheduler. Specifically, the \selectors employ hypernetworks to construct task-specific soft masks, which can be applied by globally shared representation to construct task-specific features. The task scheduler selects tasks for learning through two metrics, where the selection probability is inversely proportional to task progress (e.g., success rate) and task learning speed. Experimental results show that \ours consistently outperforms the state-of-the-art MTRL algorithms on various robotics manipulation tasks.

\end{abstract}

\begin{IEEEkeywords}
reinforcement learning; multi-task learning; knowledge sharing; task scheduler
\end{IEEEkeywords}

\section{Introduction}
Deep reinforcement learning (DRL) has been applied to solve various decision-making problems, including games \cite{mnih2015human,silver2016mastering,berner2019dota, yang2021efficient, 10.1109/ASE.2019.00077}, robot control \cite{ibarz2021train, fu2021towards}, and autonomous driving \cite{sallab2017deep,kiran2021deep, kai2020multi, pmlr-v155-zhou21a}.
Despite significant success in single-task learning, it was achieved one task at a time, with each task requiring the training of a new agent from scratch. This typically results in high memory usage, high computational cost, and, more importantly, knowledge cannot transfer between tasks during the training process, leading to low learning efficiency.
One promising way to improve learning efficiency is multi-task reinforcement learning (MTRL), which uses shared representations between a collection of related tasks.
However, MTRL is still challenging, since different tasks usually interfere during training, dramatically reducing the learning efficiency \cite{crawshaw2020multitask}. Training with many tasks at the same time, for example, using a shared network trunk and several task-specific layers may impair overall performance as compared to independent training in each task \cite{pmlr-v100-yu20a}. This phenomenon is called \emph{destructive interference}, since we do not know how the tasks will affect one another when training together in a single model.
\begin{figure}[tp]
    \centering
    \includegraphics[width=\columnwidth]{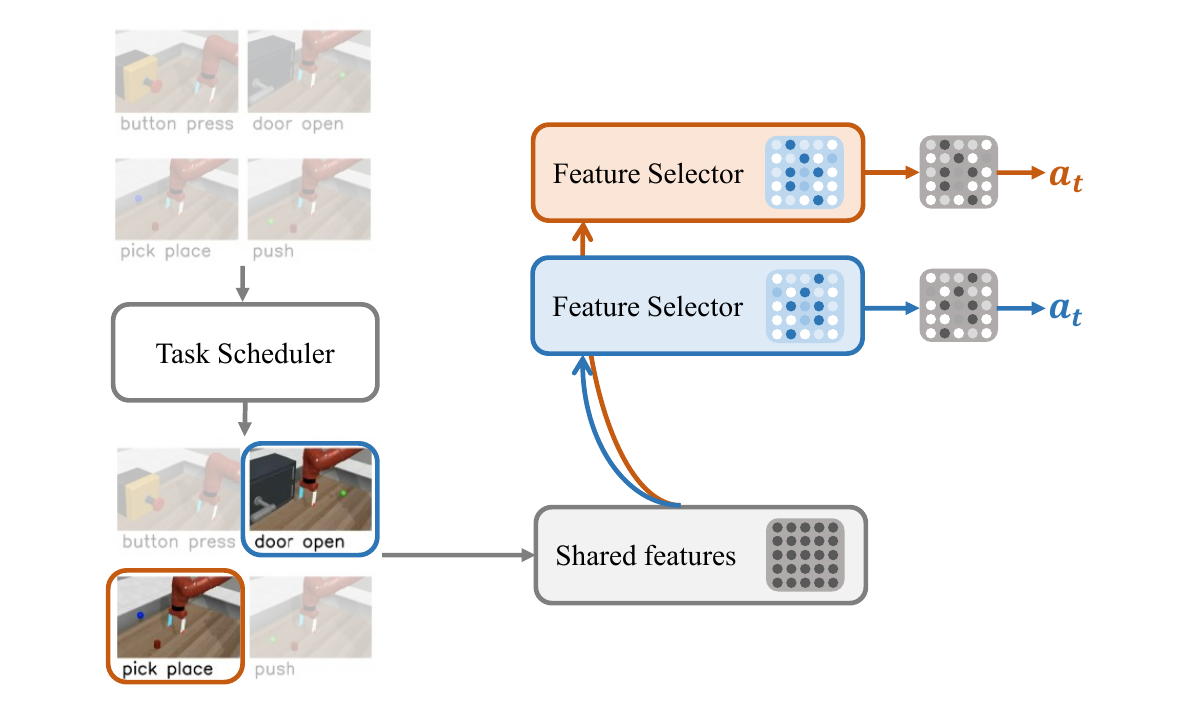}
    \caption{Overview of our framework \ours. To balance the task difficulty, the task scheduler first selects challenging tasks for the multi-task agent from the whole task set based on the performance metric. Then the \selector learns which features to be shared between tasks and which not, by filtering task-specific features from globally shared features in an end-to-end manner.}
    \label{fig:overview}
\end{figure}

One major branch of MTRL focuses on the network architecture design to resolve \emph{destructive interference}.
For example, Cross-Stitch Network \cite{misra2016crossstitch} learns static linear combinations to fuse features of different tasks. However, the magnitude of network parameters increases as the number of tasks increases, which is computationally expensive.
Multi-gate Mixture-of-Experts (MMoE) \cite{ma2018modeling} applies gating networks to combine experts (feed-forward networks) based on the input to handle task differences. However, all outputs from the experts are weighted according to the gating networks without discriminating between shared and specific ones, leading to \emph{destructive interference}.
Recently, MTAN \cite{liu2019endtoend} consists of a shared network along with an attention module for each task to obtain task-specific features. However, MTAN is mainly designed to extract image features, while our network focuses on the state input in vector form, so it is more suitable for multi-task learning on RL settings.  Furthermore, our network employs hypernetworks to construct task-specific layers, which can decouple the task context from the state and achieve better learning performance. Soft Modularization \cite{yang2020multitask} performs routing in a shared policy network that contains many modules (sub-networks) to learn different policies for different tasks.
However, the sharing mechanism is still coarse-grained due to using modules as basic sharing units.

Another branch of MTRL focuses on optimization strategies to address \emph{destructive interference}. From an optimization view, interference occurs as the existence of conflicting task gradients, which is based on the finding: when the angle between the gradient directions of two tasks is large, using one task gradient to update may reduces the performance of another task. For example, PCGrad \cite{yu2020gradient} addresses the conflicts by gradient projection. More recently, CAGrad \cite{liu2021conflict} reduces conflicting gradients between tasks by exploiting the worst local improvement of tasks. However, it requires many optimization steps during the training process, which are computationally expensive.

To solve the above problems, we propose a novel MTRL framework called \fullours. Figure \ref{fig:overview} illustrates the overall framework, which consists of two components to promote knowledge sharing between tasks.
The first component is a \selector, which employs hypernetworks and takes the task ID as input to construct a task-specific soft mask, enabling inter-task parameter sharing at the feature level, which is fine-grained.
The second component is a task scheduler, which is based on the fact that task difficulty disparities can lead to an inappropriate focus on easy tasks, slowing learning progress on challenging tasks \cite{sharma2018learning}.
In summary, our contributions are three-fold:
\begin{itemize}
    \item Our novel MTRL framework \ours comprises a \selector and a task scheduler. Those two components are designed to work in a fine-grained manner to promote knowledge flow between similar tasks.
    \item Our task scheduler efficiently schedules tasks for learning through two task metrics: task progress and task learning speed, where the selection probability is inversely proportional to task progress and task learning speed.
    \item Our framework can be easily combined with existing off-policy DRL algorithms. Experimental results show that \ours  significantly outperforms the state-of-the-art MTRL algorithms on various robotics manipulation tasks.
\end{itemize}

\section{Background}
\noindent{\textbf{Problem Settings.}}
Typically, we model the RL problem as a finite Markov Decision Process (MDP) for each task, which can be described by $(S, A, P, R, H, \gamma)$, where $S$ and $A$ are the space of states and actions. $P\left(s_{t+1} \mid s_{t}, a_{t}\right)$ denotes the state transition function. $R(s_t,a_t)$ denotes the reward function, $H$ is the horizon, and $\gamma$ represents the discount factor. The goal of the agent is to learn an optimal policy $\pi^{*}$ that maximizes the expected discounted return $R=\sum_{i=t}^{T} \gamma^{i-t} r_{i}$. 
In this paper, we follow a common MTRL setting \cite{yang2020multitask, yu2020gradient, liu2021conflict}, where a set of tasks are treated equally, each of which may have a different MDP. For example, the multi-task environment Meta-World contains a variety of object manipulation tasks such as opening a door and closing a window, 

\noindent{\textbf{Soft Actor-Critic.}} 
Soft Actor-Critic (SAC) \cite{haarnoja2018soft} is one of the most efficient off-policy algorithms, where the actor aims to accomplish the task while acting as randomly as possible. The policy $\pi_{\theta}$ is trained to maximize a trade-off between expected return and action distribution entropy:
\begin{equation*}
    L_{SAC}^{\theta}=\mathbb{E}_{\tau}\left[\min _{i=1,2} Q_{\phi_i}(s_t, a_t)-\alpha \log \pi_{\theta}\left(a_{t} \mid s_{t}\right)\right],
\end{equation*}
where $\alpha$ is the temperature parameter to maintain the entropy level of policy, controlling the stochasticity of the policy $\pi_{\theta}$.
The target value function can be calculated by the Bellman equation:
\begin{equation*}
    \begin{split}
        y\left(r, s^{\prime}\right)=r+
        \gamma\left(\min _{i=1,2} \bar{Q}_{\phi_i}\left(s^{\prime}, \tilde{a}^{\prime}\right)-\alpha \log \pi_{\theta}\left(\tilde{a}^{\prime} \mid s^{\prime}\right)\right),
    \end{split}
\end{equation*}
where $\tilde{a}^{\prime} \sim \pi_{\theta}\left(\cdot \mid s^{\prime}\right)$ and $\bar{Q}_{\phi_i}$ is target network.
The value network $Q_{\phi_i}$ can be updated using the TD error:
$L_{SAC}^{\phi_i} = \mathbb{E}_{\tau} [Q_{\phi_{i}}(s, a)-y(r, s^{\prime})]^2.$
Thus, the overall SAC optimization objective is:
\begin{equation} \label{eq:sac}
    L_{SAC} = L_{SAC}^{\theta} + L_{SAC}^{\phi_1} + L_{SAC}^{\phi_2}.
\end{equation}
Note that different tasks may have different learning statuses, we assign a separate temperature $\alpha$ for each task. 
The approach for optimizing policy and critic networks stays the same as in the standard SAC algorithm.

\begin{figure*}[htb]
    \centering
    \includegraphics[width=1\textwidth]{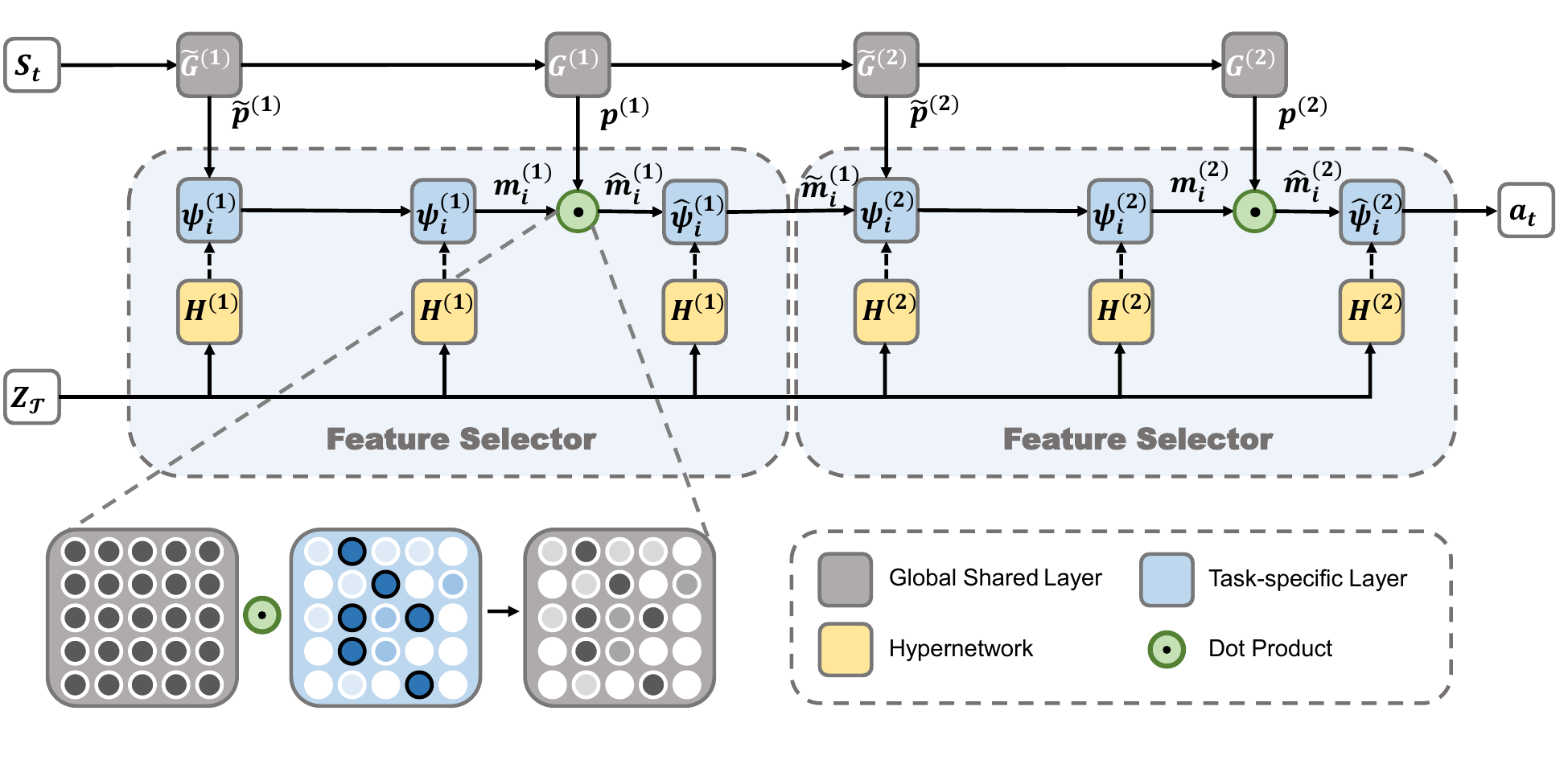} %
    \caption{Visualisation of our network architecture, showing the \shared and \selectors respectively. The role of the \selectors is to extract task-specific features from the \shared using the soft masks. The number of \selectors can be defined according to the task complexity. Here the solid line (black) denotes the data flow, and the dashed line (black) denotes the weights flow.}
    \label{fig:framework}
\end{figure*}

\section{Methodology}
In this section, we propose a MTRL framework \ours, which consists of a novel network architecture and a task scheduler.
In Section \ref{sec:architecture}, we describe the network architecture based on \selectors and hypernetworks to learn soft masks for different tasks, enabling inter-task parameter sharing at the feature level.
In Section \ref{sec:scheduler}, we introduce a scheduling mechanism where the task scheduler can focus on complex tasks, improving sample efficiency and reducing inter-task interference.
Finally, in Section \ref{sec:alg}, we describe how \ours combines with a specific DRL algorithm SAC \cite{haarnoja2018soft}.

\subsection{Task-Specific Feature Selector}\label{sec:architecture}
In this section, we describe how we design the MTRL network architecture. We propose to perform MTRL using soft masks applied to the globally shared network. 
Figure \ref{fig:framework} illustrates our proposed network architecture in detail, which consists of two parts: a \shared and several \selectors. Specifically, the \shared is used to extract the shared features from all tasks, while \selectors are designed to learn the soft masks from the \shared. The soft masks are then used to extract the task-specific features in a fine-grained manner.

Firstly, we need to generate soft masks for each task based on the task context. Previous works \cite{mcclelland1985putting, schmidhuber1992learning, sarafian2021recomposing, huang2021continual} demonstrate that using a primary feed-forward network to generate weights for a dynamic network is appropriate for this type of context-dependent function. Thus we generate soft masks using hypernetwork \cite{ha2016hypernetworks}. The hypernetwork takes the task context $z_{\mathcal{T}}$ (e.g., task ID one-hot) as input and outputs the weights of the mask generator. Then, the mask generator takes the shared features as input and outputs the soft mask $m_{i}^{(j)}$ for task $i$ at the $j^{th}$ \selector. The gradient flows through the mask generator to the weights of hypernetwork. 

We denote the output feature in the $j^{th}$ \selector of the globally shared layer $G_j^{(j)}$ as $p^{(j)}$, and denote $\tilde{p}^{(j)}$ as the output feature of the globally shared layer $\tilde{G}^{(j)}$ to generate the soft mask. The soft mask $m_{i}^{(j)}$ in the first feature selector can be computed by:
\begin{equation*}
    m_{i}^{(1)} =\sigma({\psi_{i}^{(1)}(\tilde{p}^{(1)})}),
\end{equation*}
where:
\begin{equation*}
\tilde{p}^{(1)} = \tilde{G}^{\left(1\right)}\left(S_t\right),
\end{equation*}
\begin{equation*}
    \psi_{i}^{(j)}=H^{\left(j\right)}\left(z_{\mathcal{T}}\right).
\end{equation*}
Here $h_{i}$ denotes the hypernetwork for task $i$, and $\psi_{i}$ denotes the soft mask generator. The sigmoid activation function $\sigma$ is used to control the degree of sharing. When an element in $m_{i}^{(j)}$ tends to 1, it can be assumed that this feature is shared by all tasks, while 0 tends to be task-specific.
After get the soft mask $m_{i}^{(j)}$ for task $i$, we can obtain the task-specific features $\hat{m}_{i}^{(j)}$ as follows:
\begin{equation*}
    \hat{m}_{i}^{(j)}=m_{i}^{(j)} \odot p^{(j)},
\end{equation*}
where $\odot$ denotes the dot product multiplication. Then, the task-specific features $\hat{m}_{i}^{(j)}$ will be input to the feature extractor $\hat{\psi}_{i}^{(j)}$ for passing to the next \selector:
\begin{equation*}
    m_{i}^{(j)} = \sigma({\psi_{i}^{(j)}(\textrm{Concat}(\tilde{p}^{(j)}; \hat{\psi}_{i}^{(j-1)}(\hat{m}_{i}^{(j-1)}) ))}), \quad j \geq 2
\end{equation*}
where concat represents feature concatenation. Therefore, the shared network features and soft masks can be trained together to optimize task-specific performance, while mitigating inter-task interference in a fine-grained manner.

\subsection{Task Scheduler}\label{sec:scheduler}
\begin{algorithm}[tb]
    \caption{\ours-SAC} \label{alg}
    \begin{algorithmic}[1]
        \State \textbf{Input:} initial actor network parameters $\theta$, critic network parameters $\phi_1,\phi_2$ and its target parameters $\bar{\phi_1},\bar{\phi_2}$, task set $\mathcal{T}$, task sampling distribution $\mathcal{P}$, task sample number $K$, evaluation interval $e$, replay buffer $\mathcal{D}$.
        \State Set target network parameters $\bar{\phi}_1 \leftarrow \phi_1,\bar{\phi}_2 \leftarrow \phi_2$.
        \For {$i \in {1, ..., N}$}
        \State $\mathcal{P}_i \leftarrow \frac{1}{N}$ \Comment{Uniform distribution}
        \EndFor
        \For {step $=1, 2, ..., \textrm{maximum steps}$}

        \If {$\textrm{step} \% e = 0$}
        \State \textcolor{gray}{// Time to evaluate.}
        \For {each task $t \in \mathcal{T}$}
        \State Evaluate each task $t$ and calculate the task metric $c_i^{(1)},c_i^{(2)}$. \Comment{Eq.~\eqref{eq:m1}\eqref{eq:m2}}
        \EndFor
        \State Update task sampling distribution $\mathcal{P}$. \Comment{Eq.~\eqref{eq:p}}
        \Else
        \State \textcolor{gray}{// Time to training.}
        \State Sample task subset $\mathcal{S} \sim \mathcal{P}$ ($|\mathcal{S}| = K$).

        \For {each task $t \in \mathcal{S}$}
        \State Collect trajectories for task $t$ and save to $\mathcal{D}$.
        \EndFor

        \For {minibatch $m \sim \mathcal{D}$}
        \State Update $\theta,\phi_1,\phi_2$ with $L_{SAC}$. \Comment{Eq.~\eqref{eq:sac}}
        \State Update target network $\bar{\phi_1},\bar{\phi_2}$.
        \EndFor

        \EndIf
        \EndFor
    \end{algorithmic}
\end{algorithm}

This section focuses on how our framework \ours performs task scheduling. The task scheduler selects which tasks to sample and learn during each training step. Most MTRL algorithms only consider the most straightforward method, sampling from all tasks evenly, which is typically inefficient because simple tasks converge early during the training process and should not be continuously sampled and learned. Previous work also has shown that optimized task scheduling instead of uniform sampling can significantly improve model performance \cite{bengio2009curriculum}. However, they only consider task progress rather than learning speed. Thus, the difficult task of slow learning speed is ignored to some extent, resulting in imbalanced learning.

To this end, we propose a new task scheduler concerning both task progress and learning speed. The intuition behind our task scheduler is to assign task sample probabilities based on the relative progress and learning speed between tasks: the slower the task progress and the learning speed, the more likely it is that these tasks will be sampled. We always maintain a task sampling probabilities $\mathcal{P}$ in \ours. The task sampling decision steps occur at the end of every update. We define task progress metrics as follows \cite{sharma2018learning}:
\begin{equation}\label{eq:m1}
    c_{i}^{(1)}=1-\rho_{i},
\end{equation}
where $\rho_{i}$ denotes the performance of the task $i$ (e.g., normalized total reward or success rate). Furthermore, we define the task learning speed metric as:
\begin{equation}\label{eq:m2}
    c_{i}^{(2)}=-\Delta \rho_{i},
\end{equation}
where $\Delta \rho_{i}$ denotes the increment performance of task $i$ at time interval $\Delta t$. Specifically, we increase the task’s sample probability when the learning speed is slow to focus on this task. Thus, the sample probability $\mathcal{P}_i^{(k)}$ for task $i$ is proportional to the exponential of metric $c_i^{(k)}$:
\begin{equation*}
    \mathcal{P}_i^{(k)}=\textrm{softmax}\left(\frac{c_{i}^{(k)}}{\tau}\right),
\end{equation*}
where $\tau$ is used to shape the sample distribution. In order to consider the above factors jointly, a weighted average is applied to compute the sampling probability:
\begin{equation}\label{eq:p}
    \mathcal{P}_i=\sum_{k} \alpha_{k} \mathcal{P}_i^{(k)}, \quad s.t. \sum_{k} \alpha_{k}=1
\end{equation}
where $\alpha_{k}$ represents the weight of task metric $c^{(k)}$.

\subsection{\ours-SAC}\label{sec:alg}
Our framework is easily combined with existing DRL algorithms. Algorithm \ref{alg} presents our framework \ours combined with one sota DRL algorithm SAC \cite{haarnoja2018soft}. \ours-SAC first initializes the task distribution $\mathcal{P}$, e.g., a uniform distribution (Lines 5-6). During the evaluation, we update $\mathcal{P}$ according to task metrics $c_i^{(j)}$, making it more focused on solving complex tasks (Lines 9-12). During the training, the task set $\mathcal{S}$ is first sampled according to task sampling distribution $\mathcal{P}$ (Lines 14-16). Then collect trajectories from the task set $\mathcal{S}$ in parallel and save them to the buffer $\mathcal{D}$. Finally, the agent computes the RL loss and updates the target networks (Lines 17-19).
It is worth noting that the samples taken from the buffer $\mathcal{D}$ usually contain all tasks, which can avoid catastrophic forgetting for well-learned tasks. However, because tasks are well chosen in the collection phase, the proportion of tasks sampled for update varies during training, with more complex tasks containing more samples.

\section{Experiements}
\begin{figure}[tp]
    \centering
    \includegraphics[width=0.48\textwidth]{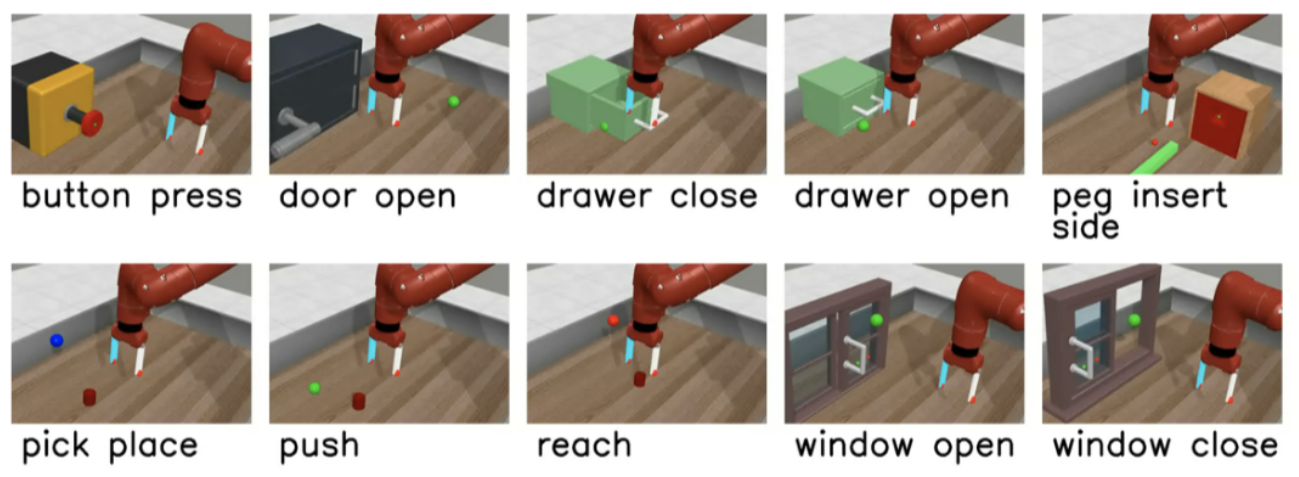}
    \caption{Illustration of the 10 robotic arm manipulation tasks on MT10.}
    \label{fig:metaworld}
\end{figure}

\subsection{Experimental Results}
\begin{figure*}[tb]
    \centering
    \subfigure[MT10-FIXED]{
        \label{fig:mt10-fixed}
        \includegraphics[width=0.24\textwidth]{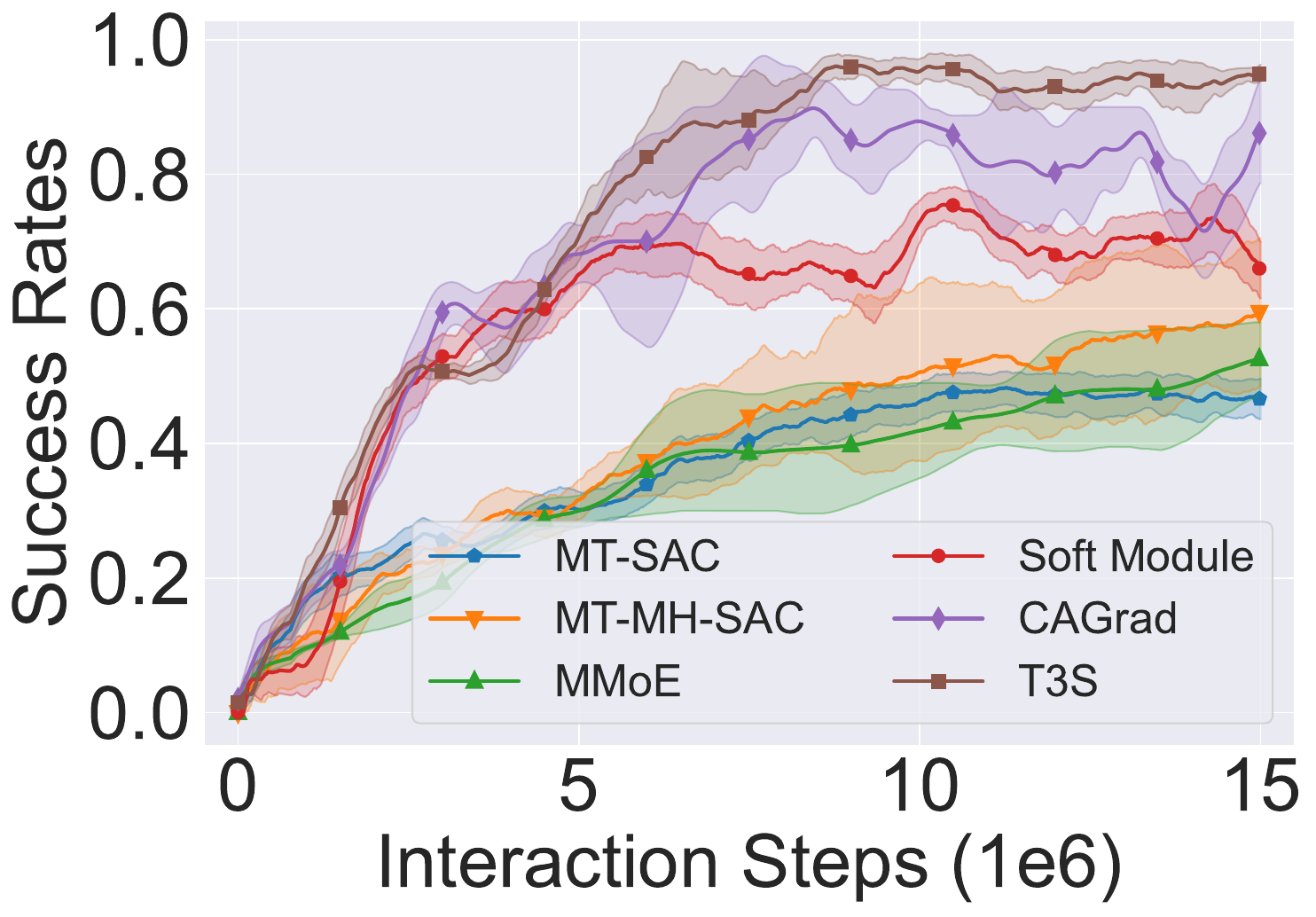}
    }
    \hspace{-4mm}
    \subfigure[MT10-RAND]{
        \label{fig:mt10-rand}
        \includegraphics[width=0.24\textwidth]{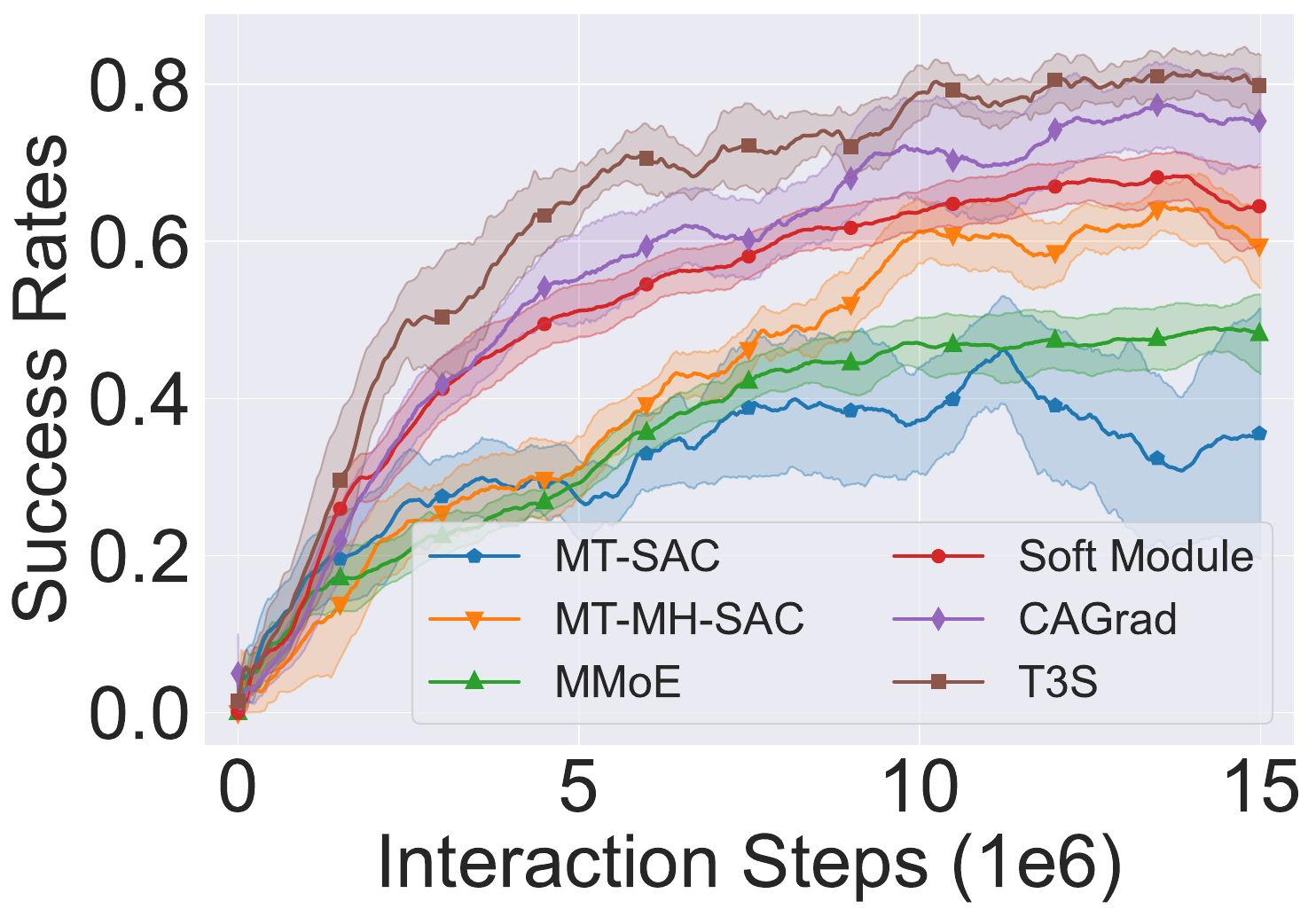}
    }
    \hspace{-4mm}
    \subfigure[MT50-FIXED]{
        \label{fig:mt50-fixed}
        \includegraphics[width=0.24\textwidth]{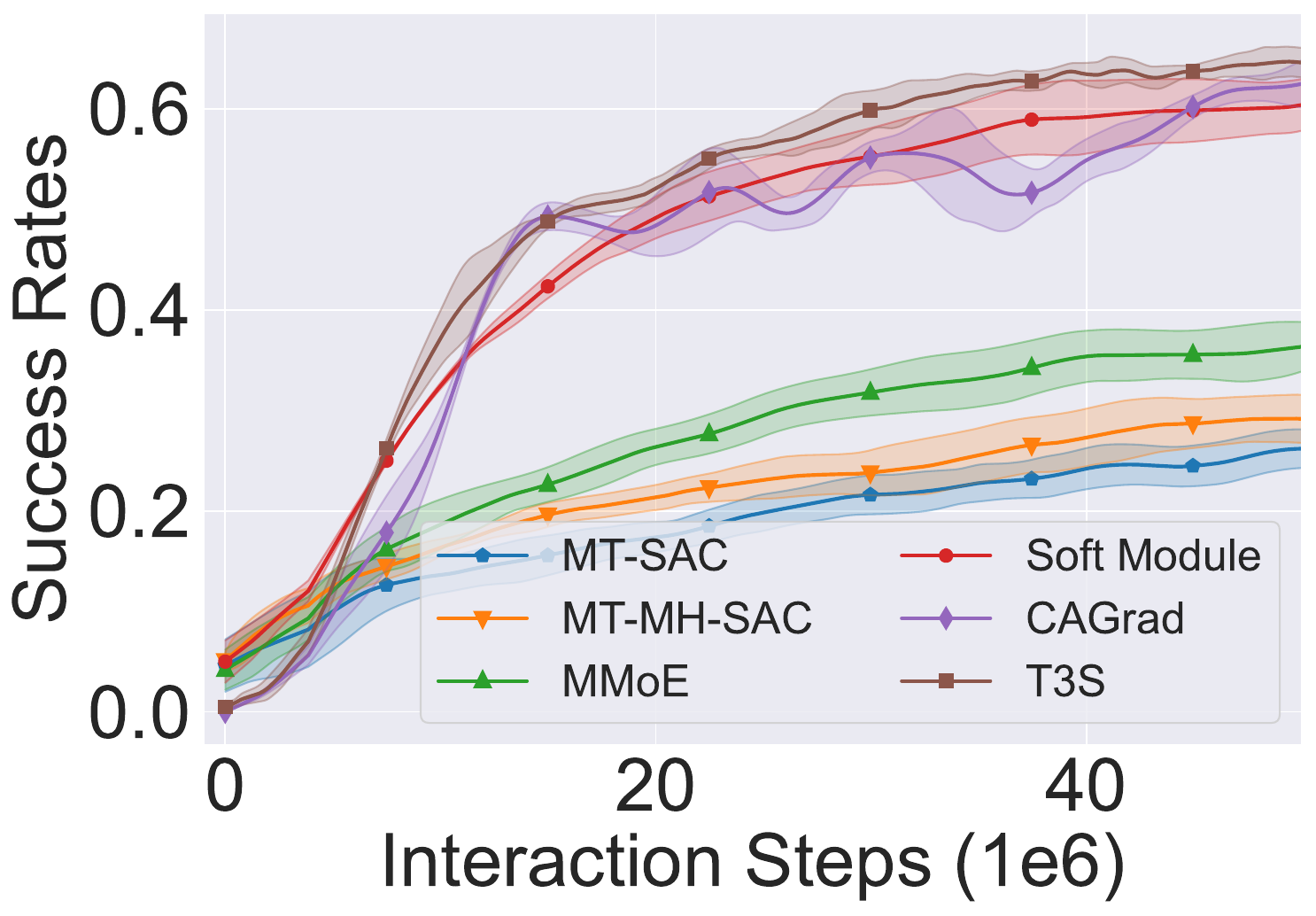}}
    \hspace{-4mm}
    \subfigure[MT50-RAND]{
        \label{fig:mt50-rand}
        \includegraphics[width=0.24\textwidth]{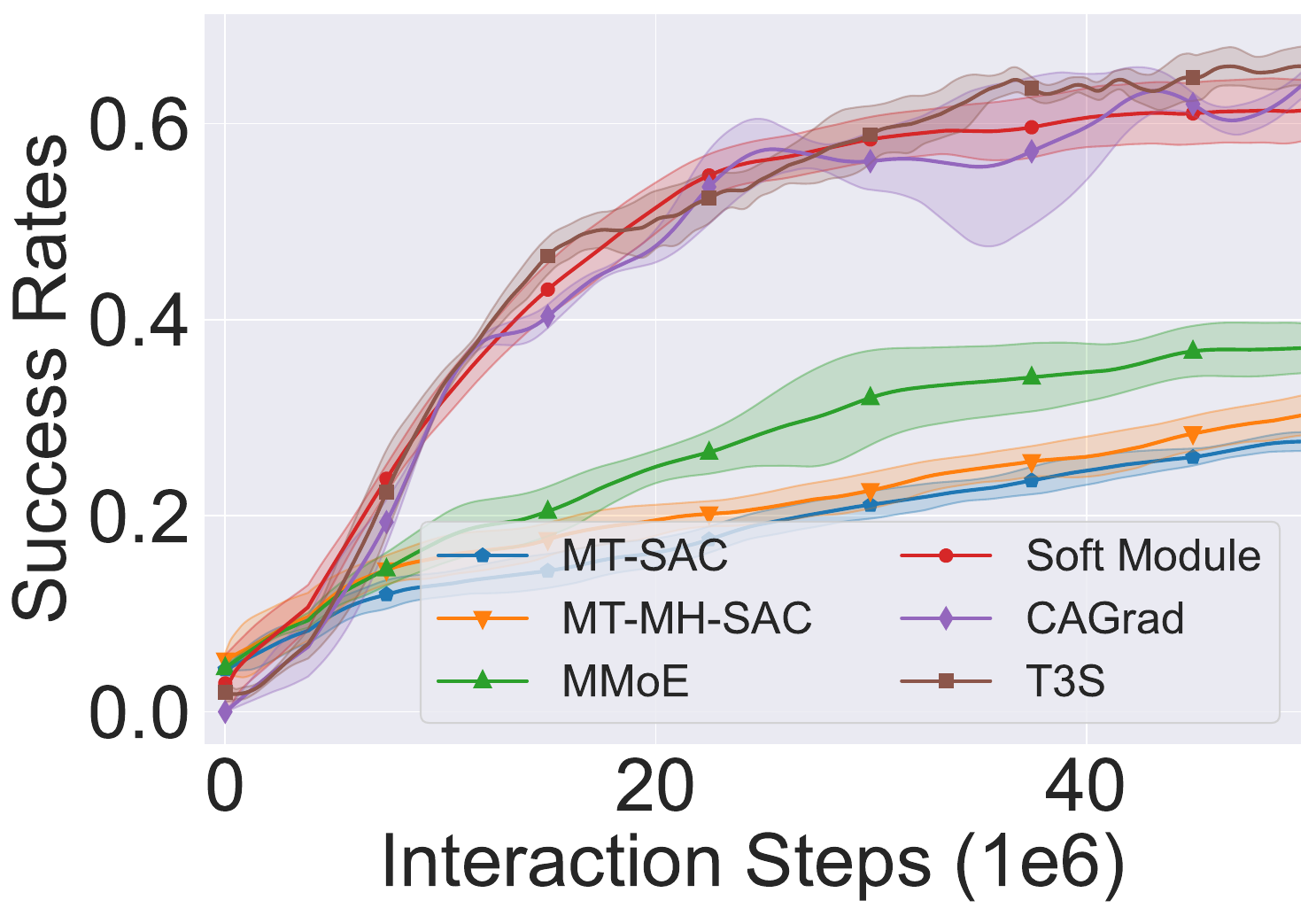}
    }

    \caption{Training curves of our proposed framework \ours and other baselines on MT10 and MT50 (shaded areas represent the standard deviation over 3 seeds). We plot the average success rate for all tasks on the y-axis, while the x-axis denotes the number of times the multi-task agent interacts with the environment. }
    \label{fig:results}
\end{figure*}

\begin{table*}[tb]
    \caption {Comparisons on success rates on all benchmarks, where success rates are averaged over the last 200,000 steps on 3 seeds.}
    \centering
    \begin{tabular}{@{}lcccc@{}}
        \toprule
        \multicolumn{1}{c}{\textbf{Method}} & \textbf{MT10-FIXED}       & \textbf{MT10-RAND}        & \textbf{MT50-FIXED}       & \textbf{MT50-RAND}        \\ \midrule
        MT-SAC\cite{pmlr-v100-yu20a}                              & 0.47 $\pm$ 0.045          & 0.36 $\pm$ 0.180          & 0.26 $\pm$ 0.019          & 0.25 $\pm$ 0.007          \\
        MT-MH-SAC\cite{pmlr-v100-yu20a}                           & 0.65 $\pm$ 0.105          & 0.61 $\pm$ 0.085          & 0.29 $\pm$ 0.023          & 0.29 $\pm$ 0.026          \\
        MMoE\cite{ma2018modeling}                                & 0.53 $\pm$ 0.045          & 0.47 $\pm$ 0.053          & 0.37 $\pm$ 0.019          & 0.36 $\pm$ 0.024          \\
        Soft Module\cite{yang2020multitask}                         & 0.67 $\pm$ 0.028          & 0.64 $\pm$ 0.047          & 0.61 $\pm$ 0.023          & 0.62 $\pm$ 0.021          \\
        CAGrad\cite{liu2021conflict}                              & 0.89 $\pm$ 0.096          & 0.75 $\pm$  0.056         & 0.63 $\pm$ 0.039          & \textbf{0.67 $\pm$ 0.029} \\
        T3S-SAC                                & \textbf{0.95 $\pm$ 0.022} & \textbf{0.80 $\pm$ 0.009} & \textbf{0.65 $\pm$ 0.005} & 0.65 $\pm$ 0.014          \\ \bottomrule[1.5pt]
    \end{tabular}

    \label{tab:results}
\end{table*}
In this section, we conduct extensive experiments to verify the effectiveness of our proposed framework \ours compared with previous multi-task methods. We introduce the environment and baselines and then compare our framework with baselines. Further, we conduct a few ablation studies to demonstrate the effectiveness of the task scheduler.

\noindent \textbf{Environments.} We evaluate our framework \ours on the MTRL representative benchmark: Meta-World \cite{pmlr-v100-yu20a}. In this environment, we need to solve many robotics continuous control and manipulation tasks with a robotic arm. In particular, the original MT10 and MT50 contain 10 and 50 robot manipulation tasks with fixed goals. We also consider more challenging settings where the tasks have random initial goals, as in \cite{yang2020multitask}. MTn-FIXED and MTn-RAND denote the environment of n tasks with fixed goals and random goals respectively. Figure \ref{fig:metaworld} shows an illustration of all the MT10 tasks.

\noindent \textbf{Baselines.} We compare \ours with representative MTRL algorithms, all baselines except MMoE are following their official source code, and MMoE is implemented following the configuration of the paper:
\begin{itemize}
    \item \textbf{Multi-task SAC (MT-SAC) }\cite{pmlr-v100-yu20a}: Using a concatenation of task ID one-hot and the state as input, and all tasks share the same network.
    \item \textbf{Multi-task multi-head SAC (MT-MH-SAC)}\cite{pmlr-v100-yu20a}: Built upon MT-SAC with separate output layers for tasks.
    \item \textbf{Multi-gate Mixture-of-Experts (MMoE)} \cite{ma2018modeling}: Sharing the expert models and training a gating network to optimize each task individually.
    \item \textbf{Soft Module} \cite{yang2020multitask}: Routing different modules in a shared model to form different policies.
    \item \textbf{CAGrad} \cite{liu2021conflict}: A gradient-based multi-task algorithm to deal with conflicting gradients.
\end{itemize}

\noindent \textbf{Experimental Setup.}  We use Adam optimizer \cite{kingma2014adam} with a learning rate of $3\times10^{-4}$ for all methods. For MT10 tasks, all methods are trained over 15 million steps with a batch size of 1280. The number of feature selectors in the T3S network architecture is 1. We empirically find that when the number of feature selectors increases, the performance will decrease on MT10. The reason is that the additional complexities from over-parameterization will result in low sample efficiency. For MT50 tasks, all methods are trained over 50 million steps with a batch size of 6400. The number of feature selectors in the T3S network architecture is 2. The weight of the task metric used in the task scheduler is 0.5, which indicates that all metrics are of equal importance.

All algorithms are trained from scratch. Table \ref{tab:results} shows the quantitative results and the success rates are averaged over the last 200,000 training steps on 3 seeds. We also plot the average success rate for all tasks of our proposed framework \ours and other baselines in Figure \ref{fig:results}.

\begin{figure}[tb]
    \centering
    \includegraphics[width=\columnwidth]{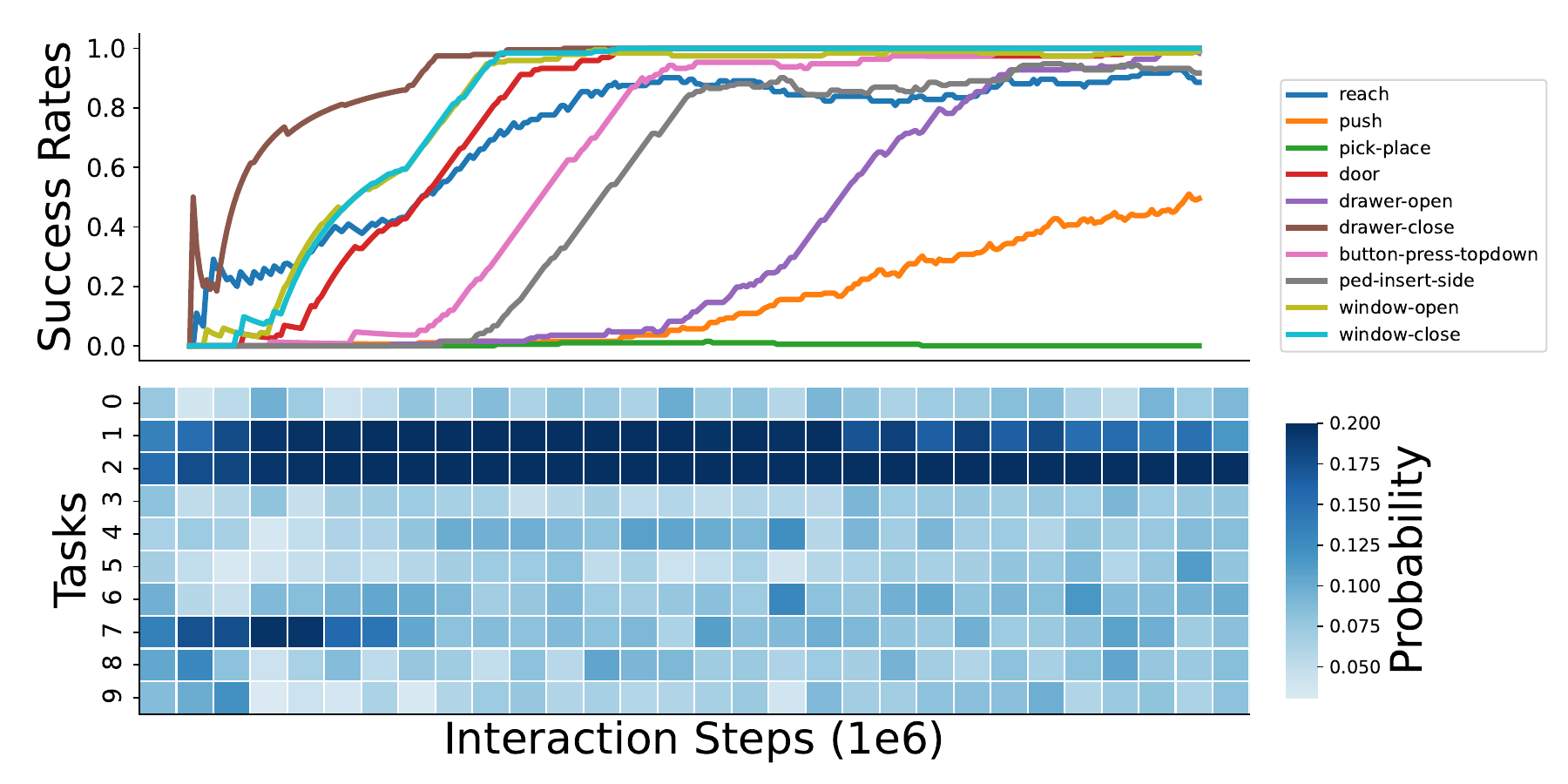}
    \caption{Sampling history of the task scheduler. (Top: line plot) The success rate for each task in MT10-RAND environment. The x-axis denotes the number of samples during training. (Bottom: square tiles) The sampling history of each task. Each tile represents the probability of sampling over a period, where darker colors denote higher probability.}
    \label{fig:history}
\end{figure}

\noindent \textbf{MT10.}  As shown in Figure \ref{fig:mt10-fixed}, our proposed framework significantly improves sample efficiency while reducing inter-task interference compared to all baselines. We further conducted experiments under the randomly generated goal setting to make the environment more challenging. A similar phenomenon can be found in Figure \ref{fig:mt10-rand} that our approach outperforms all other baselines in terms of sample efficiency and performance. This is due to the \selector used in  \ours, which automatically learns which features to be shared or task-specified in a fine-grained way by soft masks, mitigating interference between tasks. Figure \ref{fig:history} shows the sampling history of the task scheduler in MT10-RAND. It is observed that the task scheduler focuses mainly on the \emph{push} and \emph{pick-place} tasks during the entire training process, as these two tasks are the most complex. We can also observe that \emph{ped-insert-side} task performs worse than other tasks early during training, so the task scheduler is more focused on this task. When a task was gradually mastered, the task was sampled less in order to reduce unnecessary interactions.

\begin{table}[tb]
    \caption {The training time spent per update step for CAGrad and \ours.}
    \centering
    \begin{tabular}{@{}lcc@{}}
        \toprule
        \multicolumn{1}{c}{\textbf{Method}} & \textbf{MT10 Time (sec)}  & \textbf{MT50 Time (sec)}  \\ \midrule
        CAGrad                              & 10.3 $\pm$ 0.026          & 27.8 $\pm$ 0.018          \\
        Ours                                & \textbf{0.47 $\pm$ 0.032} & \textbf{1.23 $\pm$ 0.027} \\ \bottomrule
    \end{tabular}
    \label{tab:time}
\end{table}

\noindent \textbf{MT50.}  
Figures \ref{fig:mt50-fixed} and \ref{fig:mt50-rand} show the performance of \ours and baselines in MT50 with both fixed and random initial goals. We can see that \ours greatly outperforms other baselines except for CAGrad. Although \ours achieves similar performance regarding the final success rate to CAGrad, our approach exhibits some advantages in sample efficiency and stability (lower variance) due to the fine-grained sharing mechanism and the task scheduler. 
On the other hand, CAGrad is more computationally expensive than \ours. Since the time consumption of baselines other than CAGrad does not have a significant difference, we only compare T3S with the CAGrad algorithm in terms of time. Table \ref{tab:time} shows the training time spent per update step for CAGrad and \ours. Our approach is around 20x faster than CAGrad. The reason is that CAGrad requires many extra optimization steps during the training process. 

\begin{figure}[tb]
    \centering
    \includegraphics[width=0.44\textwidth]{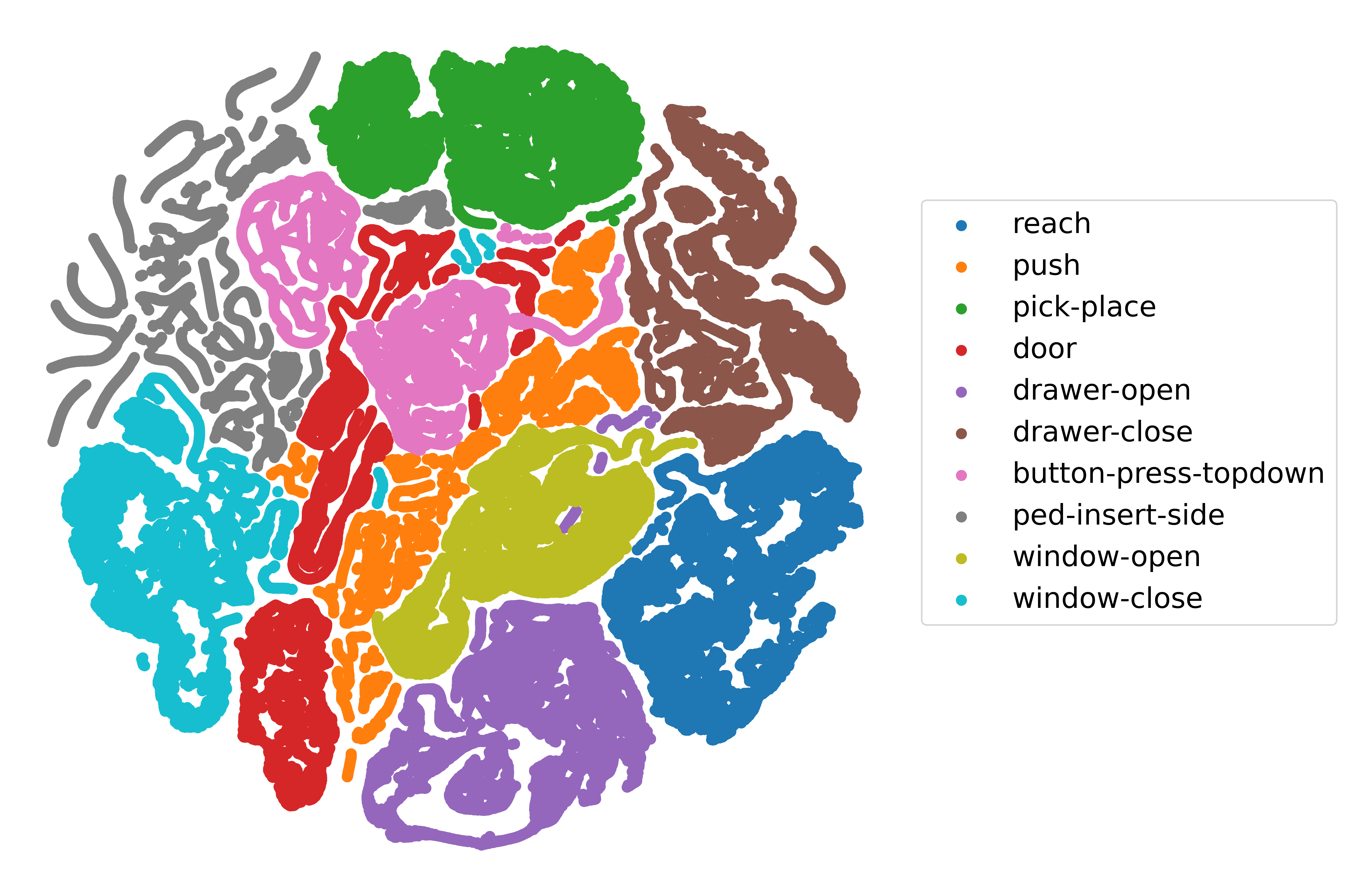}
    \caption{Visualization of soft masks for different tasks in MT10-RAND. We rollout different tasks and extract the masks from the policy. All masks are grounded in different clusters using t-SNE.}
    \label{fig:tsne}
\end{figure}

\noindent \textbf{t-SNE Visualization.} We further validate that our \ours successfully filters irrelevant information and shares similar features across tasks using soft masks. Figure \ref{fig:tsne} shows the soft masks for different tasks extract from the policy via t-SNE \cite{van2008visualizing}. We do 20 rollouts for each task, and during the rollout process, we store the masks from the policy network. After that, we use t-SNE to visualize the masks for analysis. We observe that when tasks share similar skills are closer in the plot. For example, \emph{ped-insert-side} and \emph{pick-place} both need to move an object from one place to another, and \emph{window-open} and \emph{window-close} both need to push the handle. Thus, they are close in the plot.

\subsection{Ablation Studies}
To better illustrate our framework, we further analyze the importance of the network structure and the task scheduler on MT10. The ablation experiments are set as follows:
\begin{itemize}
    \item \textbf{Ours w/o p}: Update the task sampling distribution $\mathcal{P}$ without concerning task progress.
    \item \textbf{Ours w/o ls}: Update task sampling distribution $\mathcal{P}$ without concerning task learning speed.
    \item \textbf{Ours w/o scheduler}: Using uniform sampling instead of the task scheduler to sample well-selected tasks.
    \item \textbf{MT-MH-SAC w scheduler}: Adding the task scheduler component to the MT-MH-SAC algorithm. It should be noted that when the feature selector component is removed from our framework, it is equivalent to the \emph{MT-MH-SAC w scheduler} algorithm.
    \item \textbf{Soft Module w scheduler}: Adding the task scheduler component to the Soft Module algorithm.
\end{itemize}

\begin{figure}[t]
    \centering
    \subfigure[MT10-RAND]{
        \label{fig:ablation1}
        \includegraphics[width=0.23\textwidth]{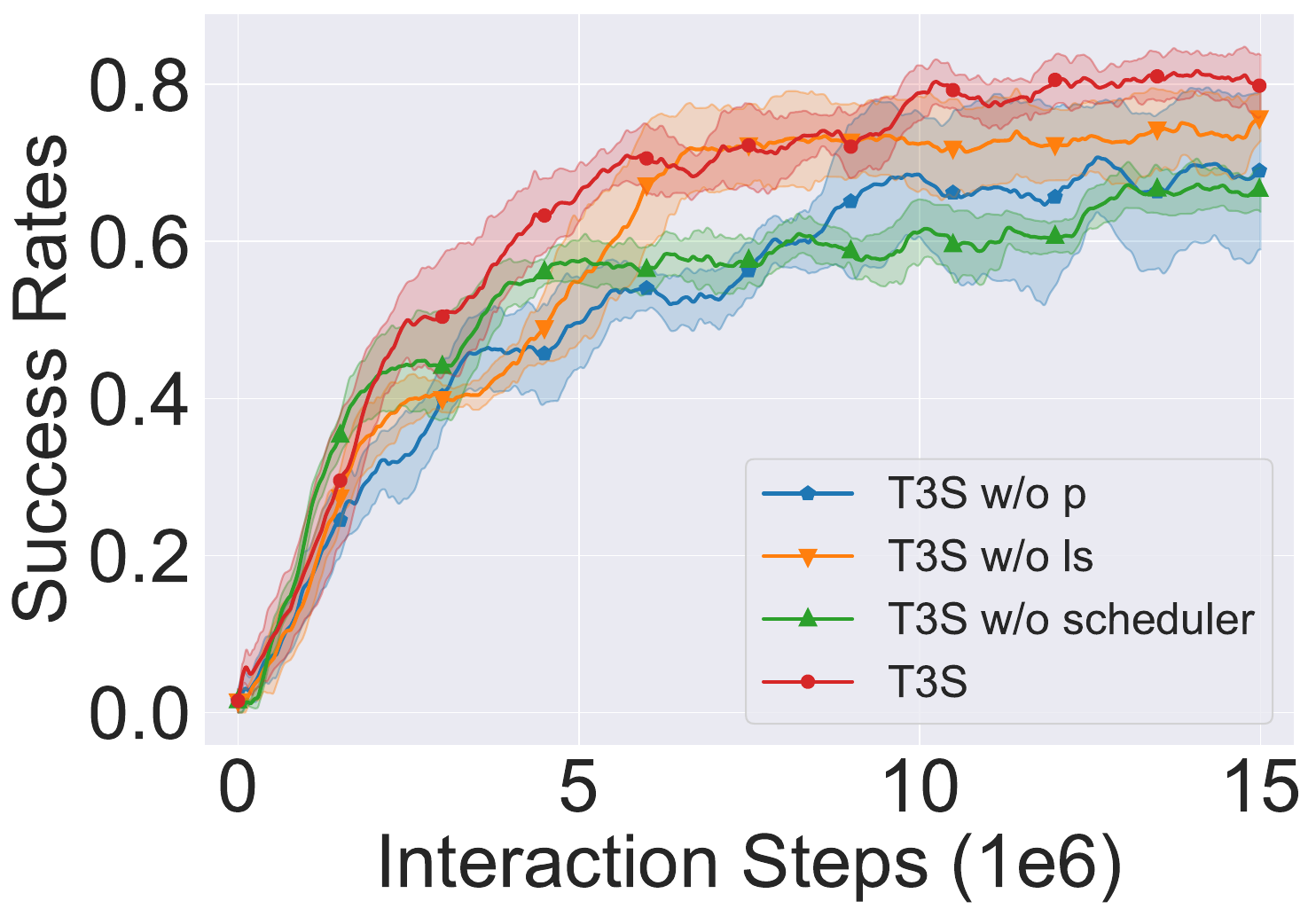}
    }
    \hspace{-4mm}
    \subfigure[MT10-FIXED]{
        \label{fig:ablation2}
        \includegraphics[width=0.23\textwidth]{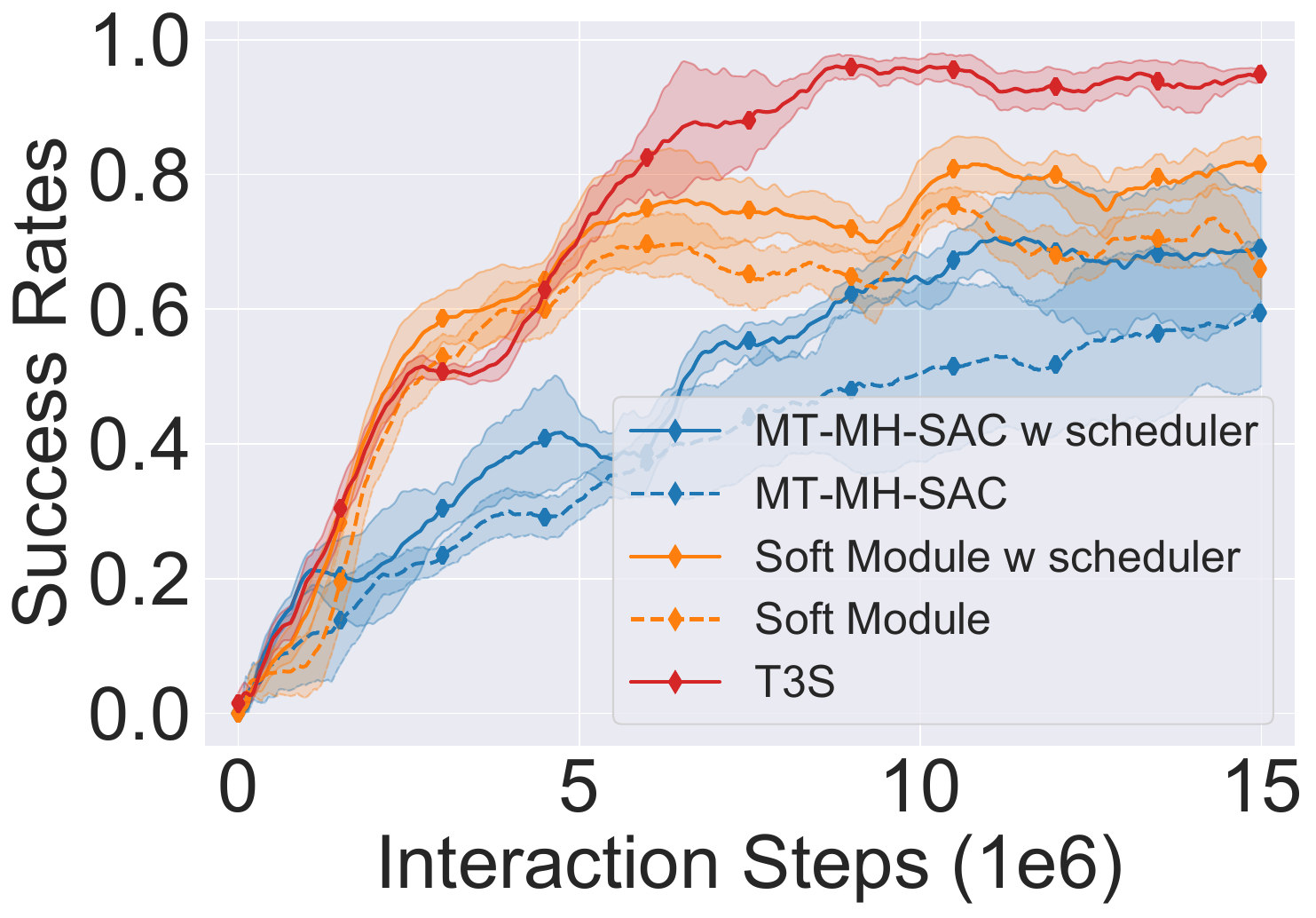}
    }
    \caption{(a) Ablation study on the task scheduler at MT10-RAND. (b) Ablation study on the feature selector at MT10-FIXED, while analysing several baselines combined with the task scheduler.}
    \label{fig:ablation}
\end{figure}

\begin{table}[t]
    \caption {Average success rate on MT10-RAND, where success rates are averaged over the last 200,000 steps on 3 seeds.}
    \centering
    \begin{tabular}{@{}lcc@{}}
        \toprule
        \multicolumn{1}{c}{\textbf{Method}} & \textbf{Success Rate}    \\ \midrule
        T3S w/o p                              & 0.76 $\pm$ 0.108                    \\
        T3S w/o ls                              & 0.78 $\pm$ 0.028                    \\
        T3S w/o scheduler                              & 0.72 $\pm$ 0.059                    \\
        T3S                                & \textbf{0.80 $\pm$ 0.009}  \\ \bottomrule
    \end{tabular}
    \label{tab:ablation1}
\end{table}

\begin{table}[t]
    \caption {Average success rate on MT10-FIXED, where success rates are averaged over the last 200,000 steps on 3 seeds.}
    \centering
    \begin{tabular}{@{}lcc@{}}
        \toprule
        \multicolumn{1}{c}{\textbf{Method}} & \textbf{Success Rate}    \\ \midrule
        MT-MH-SAC w scheduler                              & 0.73 $\pm$ 0.085                    \\
        MT-MH-SAC                              & 0.65 $\pm$ 0.105                    \\
        Soft Module w scheduler                              & 0.79 $\pm$ 0.001                    \\
        Soft Module                              & 0.67 $\pm$ 0.028                    \\
        T3S                                & \textbf{0.95 $\pm$ 0.022}  \\ \bottomrule
    \end{tabular}
    \label{tab:ablation2}
\end{table}

As shown in Figure \ref{fig:ablation1}, the performance of \ours degrades when the task scheduler is removed. The result demonstrates that our proposed task scheduler plays an essential role in MTRL. Meanwhile, as seen from \emph{T3S w/o p} (blue solid line) versus \emph{T3S w/o ls} (orange solid line), the performance gains from the task progress metric are more significant than the task learning speed metric, which indicates that we can weigh more on task progress. Figure \ref{fig:ablation2} shows the performance of the baselines when adding the task scheduler to further demonstrate the effectiveness of the scheduler module. We observe that the task scheduler improves the performance of baselines and increases the sampling efficiency compared to uniform sampling. The reason is that simpler tasks can be over-trained, while learning harder tasks may hit a plateau. Thus it is easily leading to imbalanced learning. With the help of the task scheduler, we can explicitly control the probability of each task being sampled based on the task progress and learning speed. This allows the agent to focus on the more difficult tasks, thus increasing learning efficiency. Table \ref{tab:ablation1} and \ref{tab:ablation2} show the quantitative ablation results respectively. It is worth noting that when the feature selector component is removed from the T3S network architecture, it degenerates to the \emph{MT-MH-SAC w scheduler} algorithm. Figure \ref{fig:ablation2} shows the comparison results of T3S (red solid line) and \emph{MT-MH-SAC w scheduler} (blue solid line). We can see that T3S has very impressive performance compared to \emph{MT-MH-SAC w scheduler}, which supports our view that the fine-grained sharing mechanism we proposed can mitigate inter-task interference, improving multi-task learning efficiency.

\section{Related Work}
\noindent \textbf{Multi-task Architectures.} 
Architecture design should consider which parameters of the model should be shared among tasks and which parameters should be task-specific. Thus how to balance the task-sharing and task-specific parameters is essential.
Researchers have made some progress in multi-task network architecture \cite{misra2016crossstitch, fernando2017pathnet, rosenbaum2018routing, ma2018modeling, mallya2018packnet, liu2019endtoend, sun2020adashare, tang2020progressive, yang2020multitask}.
For example, Multi-gate Mixture-of-Experts (MMoE) \cite{ma2018modeling} solves multi-task learning by sharing experts (feed-forward networks) across all tasks and training a gating network to optimize each task independently. However, the outputs of all experts are weighted without distinguishing between task-shared and task-specific ones, which can easily lead to \emph{destructive interference}.
Routing Network \cite{rosenbaum2018routing} propose a routing network and a set of function blocks, where different function blocks are dynamically combined for each input by the routing network. However, they use RL to train the routing policy. Due to the temporal nature of RL, jointly training a control policy and a routing policy may suffer from exponentially variance in policy gradient, resulting in training instability.
More recently, MTAN \cite{liu2019endtoend} is made up of a single shared network that includes a global feature pool and a soft-attention module for each task. However, MTAN is mainly designed to extract image features, while our network focuses on the state input in vector form. Soft Modularization \cite{yang2020multitask}, instead of selecting routes directly for each task as in Routing Network \cite{rosenbaum2018routing}, they use soft modules (sub-networks) to combine all possible routes. Although Soft Modularization is a successful approach for multi-task reinforcement learning, it still faces the problem of negative transfer due to its treat module as the basic sharing unit. By contrast, our network allows different tasks to be shared at the parameter level, which is more fine-grained and thus significantly mitigates interference between tasks.

\noindent \textbf{Multi-task Optimization Strategies.}
Optimization strategies are designed to handle the balance of all tasks. The difficulty of each task varies; thus, the agent should prioritize the more complex tasks. Besides, the \emph{destructive interference} phenomenon can be demonstrated from an optimization view by conflicting gradients \cite{yu2020gradient}. When the gradients of two tasks point in opposite directions, following the gradient of one task may reduces the performance of another task.
Actor-Mimic \cite{parisotto2016actormimic} and Policy Distillation \cite{rusu2016policy} both use knowledge distillation for multi-task reinforcement learning. Their insights are similar: for each task, train a task-specific teacher policy and then use policy distillation to train a single student policy to imitate the outputs of the task-specific teacher policies. Although this approach achieves competitive results compared with single-task learning, it requires separate well-training teacher policies and an extra distillation phase. Furthermore, knowledge cannot be shared during the teacher training phase.
More recently, researchers propose explicitly avoiding conflicting gradients from different tasks \cite{yu2020gradient, chen2020just, liu2021conflict}. For example, PCGrad \cite{yu2020gradient} projects the gradient of one task onto the normal plane of the gradient of any other conflicting task. CAGrad \cite{liu2021conflict} automatically balances different task objectives and has proven to converge to Pareto optimal solutions. However, they still require many optimization computations during the training process, which is computationally expensive. 

\section{Conclusion and Future Work}
We propose a novel MTRL framework called \fullours enabling efficient knowledge transfer between tasks.
\ours is composed of two main components: a \selector and a task scheduler.
Specifically, \selectors employ hypernetworks and take the task ID as input to construct soft masks, automatically choosing which features to share and which to be task-specific in a fine-grained manner.
Then we design the task scheduler, which efficiently schedules tasks for learning through two task scheduling metrics where the selection probability is inversely proportional to task progress and task learning speed.
In our experiments, we evaluate our proposed approach on various robotics manipulation tasks, outperforming several state-of-the-art MTRL baselines. We also perform a visual analysis via t-SNE to verify the task distinction capability of soft masks.
For limitations, our work is not done on real robots. In addition, the weights of each metric in the task scheduler are set manually.
In future work, we would like to design a more effective task scheduler and extend our framework \ours to real robots.

\bibliographystyle{IEEEtran}
\bibliography{root-n}{}

\end{document}